\documentclass[11pt]{article}
\usepackage{paclic32}
\usepackage[whole]{bxcjkjatype}
\usepackage{times}
\usepackage{latexsym}
\usepackage{amsmath, amsfonts}
\usepackage{booktabs}
\usepackage{multirow}
\usepackage{url}
\usepackage{xcolor}

\usepackage[pdftex]{graphicx}
\usepackage{lingmacros}
\usepackage{pxrubrica}
\usepackage{comment}
\usepackage{tabularx}

\newcommand{\bvec}[1]{\mbox{\boldmath $#1$}}

\title{Suspicious News Detection Using Micro Blog Text}

\author{
Tsubasa Tagami$^1$，Hiroki Ouchi$^2$，Hiroki Asano$^3$，Kazuaki Hanawa$^4$ \\ 
{\bf Kaori Uchiyama$^5$，Kaito Suzuki$^6$，Kentaro Inui$^7$，Atsushi Komiya$^8$} \\
{\bf Atsuo Fujimura$^9$，Hitofumi Yanai$^{10}$，Ryo Yamashita$^{11}$，Akinori Machino$^{12}$} \\
$^{1,2,3,4,5,6,7}$Tohoku University，$^{2,3,7}$RIKEN，$^{8,9}$SmartNews, Inc.，$^{10}$FactCheck Initiative Japan \\
$^{11}$Watchdog for Accuracy inNews-reporting, Japan，$^{12}$Hi-Ether Japan \\
\{$^1$tagami，$^3$asano，$^4$hanawa，$^7$inui\}@ecei.tohoku.ac.jp，$^2$hiroki.ouchi@riken.jp \\
\{$^5$kaori.uchiyama.r1，$^6$kaito.suzuki.p3\}@dc.tohoku.ac.jp \\
\{$^8$atsushi.komiya，$^9$atsuo.fujimura\}@smartnews.com \\
$^{10}$yanai@fij.info，$^{11}$rymst@gohoo.org，$^{12}$akinori.machino@hi-ether.org
}

\date{}

\begin{document}
\maketitle
\begin{abstract}
  We present a new task, {\it suspicious news detection} using micro blog text.
  This task aims to support human experts to detect suspicious news articles to be verified, which is costly but a crucial step before verifying the truthfulness of the articles.
Specifically, in this task, given a set of posts on SNS referring to a news article, the goal is to judge whether the article is suspicious or not.
  For this task, we create a publicly available dataset in Japanese and provide benchmark results by using several basic machine learning techniques.
  Experimental results show that our models can reduce the cost of manual fact-checking process.
\end{abstract}

\section{Introduction}
\label{sec:intro}

Fake news is a news article that is intentionally false and could mislead readers \cite{shu2017fake}.
The spread of fake news has a negative impact on our society and the news industry.
For this reason, fake news detection and fact-checking are getting more attention.
\\

\noindent
{\bf Problematic Issue.} \hspace{0.1cm}
One problematic issue of fake news detection is that human fact-checking experts cannot keep up with the amount of misinformation generated every day.
Fact-checking requires advanced research techniques and is intellectually demanding.
It takes about one day to fact-check a typical article and write a report to persuade readers whether it was true, false or somewhere in between \cite{hassan2015quest}.\\

\noindent
{\bf Existing Approach.} \hspace{0.1cm}
As a solution to the problem, various techniques and computational models for automatic fact-checking or fake news detection have been proposed \cite{Vlachos2014FactCT,wang:17,Hanselowski2018ARA}.
However, in practice, current computational models for automatic fake news detection cannot be used yet now due to the performance limitation.
Thus, at the present, manual or partially automatic verification is a practical solution.\\

\noindent
{\bf Our Approach.} \hspace{0.1cm}
To mitigate the problem, we aim to automate {\it suspicious news detection}.
Specifically, we develop computational models for detecting suspicious news articles to be verified by human experts.
We assume human-machine hybrid systems, in which suspicious articles are detected and sent to human experts and they verify the articles.

Our motivation of this approach is to remedy the time-consuming step to find articles to check.
Journalists have to spend hours going through a variety of investigations to identify claims (or articles) they will verify \cite{hassan2015quest}.
By automatically detecting suspicious articles, we can expect to reduce the manual cost.
\\

\noindent
{\bf Our Task.} \hspace{0.1cm}
We formalize suspicious news detection as a task.
Specifically, in this task, given a set of posts on SNS that refer to a news article, the goal is to judge whether the article is suspicious or not.
The reason of using posts on SNS is that some of them cast suspicion on the article and can be regarded as useful and reasonable resources for suspicious news detection.

This task distinguishes our work from previous work.
In previous work, the main goal is to assess the truthfulness of a pre-defined input claim (or article).
This means that it is assumed that the input claim is given in advance \cite{wang:17}.
As mentioned above, in real-world situations, we have to select the claims to be verified from a vast amount of texts.
Thus, the automation of this procedure is desired for practical fact verification.\\

\noindent
{\bf Our Dataset.} \hspace{0.1cm}
For the task, we create a Japanese suspicious news detection dataset.
On the dataset, we provide benchmark results of several models based on basic machine learning techniques.
Experimental results demonstrate that the computational models can reduce about 50\% manual cost of detecting suspicious news articles.\\

\noindent
{\bf Our Contributions.} \hspace{0.1cm}
To summarize, our main contributions are as follows,
\begin{itemize}
\item We introduce and formalize a new task, {\it suspicious news detection} using posts on SNS.
\item We create a Japanese suspicious news detection dataset, which is publicly available.\footnote{https://github.com/t-tagami/Suspicious-News-Detection}
\item We provide benchmark results on the dataset by using several basic machine learning techniques.
\end{itemize}

\section{Related Work}
\label{sec:rwork}

This section describes previous studies that tackle fake news detection.
We firstly overview basic task settings of fake news detection.
Then, we discuss several studies that share similar motivations with ours and deal with fake news detection on social media.

\subsection{Task Settings of Fake News Detection}
Typically, fake news detection or fact-checking is defined and solved as binary prediction \cite{PrezRosas2017AutomaticDO,Volkova2017SeparatingFF,Gilda2017EvaluatingML} or multi-class classification \cite{wang:17,Hassan2017TowardAF}.
In this setting, given an input text $x$, the goal is to predict an appropriate class label $y \in \mathcal{Y}$.
The input text $x$ can be a sentence (e.g., news headline, claim or statement) or document (e.g., news article or some passage).
The class labels $\mathcal{Y}$ can be binary values or multi-class labels.

One example of this task is the one defined and introduced by the pioneering work, \newcite{Vlachos2014FactCT}.
Given an input claim $x$, the goal is to predict a label $y$ from the five labels, $\mathcal{Y} = \{$\textsc{True}, \textsc{MostlyTrue}, \textsc{HalfTrue}, \textsc{MostlyFalse}, \textsc{False}$\}$.

Another example is a major shared task, Fake News Challenge.
In this task, given a headline and body text of a news article, the goal is to classify the stance of the body text relative to the claim made in the headline into one of four categories, $\mathcal{Y} = \{$\textsc{Agrees}, \textsc{Disagrees}, \textsc{Discusses}, \textsc{Unrelated}$\}$.
A lot of studies have tackled this task and improved the computational models for it. \cite{Thorne2017FakeND,Tacchini2017SomeLI,Riedel2017ASB,Zeng2017NeuralSD,Pfohl2017StanceDF,Bhatt2017OnTB,Hanselowski2018ARA}.


One limitation of the mentioned settings is that the input text is predefined.
In real-world situations, we have to select the text to be verified from a vast amount of texts generated every day.

Assuming such real-world situations, \newcite{Hassan2017TowardAF} aimed to detect important factual claims in political discourses.
They collected textual speeches of U.S. presidential candidates and annotated them with one of the three labels, $\mathcal{Y} = \{$\textsc{Non-Factual Sentence}, \textsc{Unimportant Factual Sentence}, \textsc{Check-Worthy Factual Sentence}$\}$.
There is a similarity between their work and ours.
One main difference is that while they judge whether the target political speech is check-worthy or not, we judge the degree of the suspiciousness of the target article from the posts on SNS referreing to the article.

\subsection{Fake News Detection on Social Media}
We aim to detect suspicious news using information on social media.
There is a line of previous studies that share a similar motivation with our work.

\subsection*{Fake News Detection Using Crowd Signals}
One major line of studies on fake news detection on social media leveraged crowd signals \cite{Zhang2018FakeND,Kim2018LeveragingTC,Castillo2011InformationCO,Liu2016ICEIC,Tacchini2017SomeLI}.

\newcite{Tschiatschek2017DetectingFN} aimed to minimize
the spread of misinformation by leveraging user's flag activity.
In some major SNS, such as Facebook and Twitter, users can flag a text (or story) as misinformation.
If the story receives enough flags, it is directed to a coalition of third-party fact-checking organizations, such as Snoops\footnote{http://www.snopes.com} or FactCheck\footnote{http://www.factcheck.org}.
To detect suspicious news articles and stop the propagation of fake news in the network, \newcite{Tschiatschek2017DetectingFN} used the flags as a clue.
\newcite{Kim2018LeveragingTC} also aimed to stop the spread of misinformation by leveraging user's flags. 


\subsection*{Fake News Detection Using Textual Information}
Another line of studies on fake news detection on social media effectively used textual information \cite{Mitra2015CREDBANKAL,wang:17,Tacchini2017SomeLI,PrezRosas2017AutomaticDO,Long2017FakeND,Vo2018TheRO,Yang2018TICNNCN}.


In particular, \newcite{Volkova2017SeparatingFF} is similar to our work.
They built a computational model to judge whether a news article on social media is suspicious or verified.
Also, if it is suspicious news, they classify it to one of the classes, satire, hoaxes, clickbait and propaganda.
One main difference is that while the main goal of their task is to classify the input text, our goal is to detect suspicious news articles using SNS posts.

\section{Tasks}
\label{sec:task}

\begin{figure*}[t]
\begin{center}
\includegraphics[width=\linewidth]{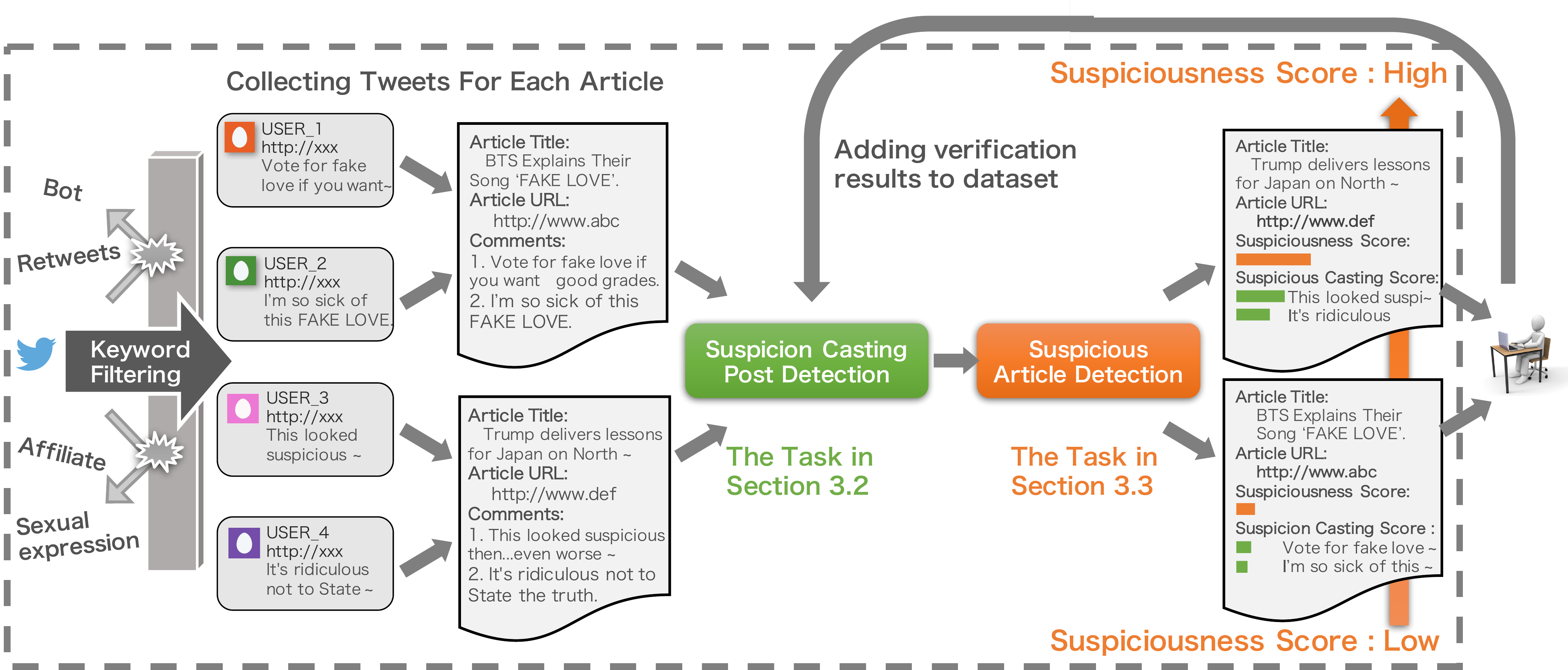}
\caption{Overall architecture of our system.}
\label{fig:system}
\end{center}
\end{figure*}

\noindent
Our main objective is to detect suspicious news articles to be verified.
In this section, we firstly explain our motivation in Section\ \ref{sec:motiv} and our system that we assume in Section\ \ref{sec:system}.
Then, we propose and formalize the two tasks, (i) {\it suspicion casting post detection} in Section\ \ref{sec:scp} and (ii) {\it suspicious article detection} in Section\ \ref{sec:sa}.

\subsection{Motivative Situation}
\label{sec:motiv}
One example of fake news detection or fact-checking in the real-world situations is the activity of Watchdog for Accuracy in News-reporting, Japan (WANJ)\footnote{http://wanj.or.jp/}, Nonprofit Organization (NPO) in Japan.
They verify news articles following the three manual steps.
\begin{enumerate}
\setlength{\parskip}{0cm}
\setlength{\itemsep}{0cm}
\item Collect the posts on SNS that refer to news articles and select only the posts that cast suspicion on the articles.
\item Select suspicious articles to be verified by taking into account the content of each collected post and the importance of the articles.
\item Verify the content of each article, and if necessary, report the investigation result.
\end{enumerate}

\noindent
In the first step, they collect and select only the SNS posts that cast suspicion on news articles.
We call them {\it suspicion casting posts} (SCP).
Based on the selected SCP, in the second and third steps, the articles to be verified are selected, and the contents are actually verified by some human experts.

All these steps are time-consuming and intellectually demanding.
Although full automation of them is ideal, it is not realistic at present due to the low performance for fact verification.
Thus, in this work, we aim to realize partial automation to support human fact-checking experts.\\

\noindent
{\bf What We Want to Do.} \hspace{0.1cm}
We aim to automate suspicious article detection by leveraging SCP information.
It is costly to collect only SCP from a vast amount of SNS posts generated every day.
Not only time-consuming, it is sometimes challenging for computational models to tell SCP from others.
Consider the follwoing two posts.
\begin{description}
\setlength{\parskip}{0cm}
\setlength{\itemsep}{0cm}
\small
\item[(a)] この記事は誤報では？千代田区も路上喫煙はダメで過料が科されているはずです！
\\ This article denotes misinformation, doesn't it?
If you had smoked on the street, you should have been fined in Chiyoda Ward!
\item[(b)] 本当に信じられない。嘘であって欲しい。言葉が見つからないけどご冥福をお祈りします！
\\ I really can not believe it.
I wish it were a lie.
I'm lost for words, but I'll send my prayers!
\end{description}

\noindent
While the post (a) casts suspicion on the article, the post (b) just mentions personal impression on it.
Acctually, only a few of the total SCP candidates are true SCP, which means that SCP detection is a heavy burden to human experts.

We develop computational models for SCP detection, and by using the resutls, we rank suspicous articles.
We assume that the suspicious articles are sent to and verified by human experts in order of suspiciousness scores.
In the following subsection, we describe the system that we assume.

\subsection{Human-Machine Hybrid System}
\label{sec:system}
Our system integrates computational models with human fact-checking experts.
Figure\ \ref{fig:system} illustrates the overall architecture of our system.
This system consists of the five components.
\begin{enumerate}
\setlength{\parskip}{0cm}
\setlength{\itemsep}{0cm}
\item Filtering Component: To collect and filter the posts on SNS referring to news articles.
\item Arranging Component: To arrange and put together the posts referring to the same article.
\item Scoring Component: To detect the posts that cast suspicion on the article and score the suspiciousness.
\item Ranking Component: To rank the articles based on the suspiciousness scores of each post.
\item Verification Component: To verify the articles by human experts.
\end{enumerate}

\noindent
For the third component, we build a scoring model by tackling a binary prediction task, SCP detection in Section\ \ref{sec:scp}.
In this task, given a post, the goal is to judge whether the post is SCP or not.
For the fourth component, we score and rank articles based on the SCP.
We define a ranking task for it, suspicious article detection in Section\ \ref{sec:sa}.
In the following subsections, we describe the task settings in detail.

\subsection{Suspicion Casting Post Detection}
\label{sec:scp}

As the example posts in Section\ \ref{sec:motiv} show, one challenge of detecting suspicion casting posts (SCP) is that a lot of posts referring to an article do not cast suspicion and just mention personal impression on the article.
Thus, a key to detecting SCP is how to capture linguistic expressions related to the truthfulness of articles.

\subsection*{Formal Setting}
Given a post $x = (w_1, \cdots, w_T)$ that consists of $T$ words and refers to an article $a \in \mathcal{A}$, the goal is to judge whether the post casts suspicion on the article or not.
\begin{align*}
\textsc{Input}&: x = (w_1, \cdots, w_T) \\
\textsc{Output}&: y \in \{ 0, 1 \}
\end{align*}

\noindent
$y$ is a binary value, i.e., $1$ represents that the post $x$ is SCP and $0$ otherwise.

\subsection*{Evaluation}
To evaluate the performance for this task, we use precision, recall and F1 scores.
If the prediction $\hat{y}$ matches with the gound-truth $y$, we regarded it as correct.

\subsection{Suspicious Article Detection}
\label{sec:sa}

\subsection*{Formal Setting}
Given an article $a$ and $N^{(a)}$ posts referring to the article $X^{(a)} = \{ x^{(a)}_i \}^{N^{(a)}}_{i=0}$, the goal is to judge whether the article is suspicious or not.
\begin{align*}
\textsc{Input}&: X^{(a)} = \{ x^{(a)}_i \}^{N^{(a)}}_{i=0} \\
\textsc{Output}&: y^{(a)} \in \{ 0, 1 \}
\end{align*}

\noindent
$x^{(a)}_i$ is each post, and $y^{(a)}$ is a binary value, i.e., $1$ represents the article is suspicious and $0$ otherwise.

\subsection*{Evaluation}
Not only precision, recall and F1 scores, we evaluate the performance using a ranking criterion, Recall@$K$.
In this work, since we assume that we send articles to human fact-checking experts in order of the suspiciousness scores, Recall@$K$ is suitable for evaluating the ability of models to properly rank the suspicious articles.


Specifically, Recall@$K$ evaluates the propotion of the correct suspicious articles in the top-$K$ ranked ones,
\begin{align*}
Recall@K = \frac{1}{|T|}\sum_{1 \leq i \leq K} b_{i} \:\:,
\end{align*}

\noindent
where $T$ is the number of the total articles in the test set, and $b_i$ is a binary value, i.e., $1$ if the $i$-th ranked article is suspicious and $0$ otherwise.

\section{Methods}
\label{sec:method}

This section describes our methods for the two tasks formalized in the previous section.

\subsection*{Suspicion Casting Post Prediction}
For SCP detection, we can simply predict $y$ based on a binary prediction approach,
\begin{align}
\label{eq:bp}
\text{P}_\theta(y=1 | x) = f_\theta(x) \:\:.
\end{align}

\noindent
$y = 1$ represents the post $x$ is SCP and $0$ otherwise.
Function $f_\theta$ with the parameters $\theta$ can be arbitrarily defined.
In this paper, as the function $f_\theta$, we use several models described in Section\ \ref{sec:model}.

To train the model parameters $\theta$, we use the binary cross-entropy loss function,
\begin{align}
\label{eq:loss}
\mathcal{L}(\theta) & = - \sum^N_{i=1}{\ell_i} \:\:,\\
\ell_i & = \log \: \text{P}_\theta(y=1 | x) + \log \: (1 - \text{P}_\theta(y=1 | \bvec{x})) \:\:.\nonumber
\end{align}

\subsection*{Suspicious Article Prediction}
For suspicious articles detection, we predict $y^{(a)}$ based on the SCP prediction score of each post.
We firstly score each of the posts $x^{(a)} \in X^{(a)}$ referring to the article $a$.
Then we use the highest score among them as the score of $y^{(a)}$.
Specifically, we calculate the score of $y^{(a)}$ as follows,
\begin{align}
\label{eq:tbv}
\textsc{Score}(y^{(a)}) = \max_{x^{(a)} \in X^{(a)}} \text{P}_\theta(y=1 | x^{(a)}) \:\:.
\end{align}

\noindent
Here, the SCP probability $\text{P}_\theta(y=1 | x^{(a)})$ can be calculated in the same way as Eq.\ \ref{eq:bp}.
We determine that the article $a$ is suspicious, i.e., $y^{(a)} = 1$, if $\textsc{Score}(y^{(a)})$ is greater than $0.5$.
The parameters $\theta$ are optimized by using the same loss function as the one for SCP prediction (Eq.\ \ref{eq:loss}).

\section{Datasets}
\label{sec:dataset}

This section describes the procedure of our dataset creation.
We created the two datasets, the one for suspicion casting post (SCP) detection and the other for suspicious article (SA) detection.
Note that these two datasets are independent sets of posts, which means that they do not share the same posts with each other.
In the following subsections, we explain the procedures in detail.

\subsection{Dataset for Suspicion Casting Post Detection}
\label{sec:data-vns}

First, we collected the posts on Twitter including the URL of a news article.
Of these posts, we left only the posts that have the potential to cast suspicion by using specific keywords, such as {\it misinformation}, {\it fabrication} and {\it untrue}.
In this work, we adopted the list of the keywords that is actually used for fact-checking by FIJ\footnote{http://fij.info/}, the third-party fact-checking organization in Japan.
If the post contains any key words in the list, we regarded it as a candidate post and added it to the dataset.

Second, we preprocessed the collected posts.
We want to leave only the comment part of a post except for some noises, such as hashtags, mentions and title of news articles.
Thus, we removed the article title, URL and hashtags from posts.
As a result, we obtained only the comment part other than noise from the original post.

Finally, to each collected post, we annotated $1$ if the post casts suspicion and $-1$ otherwise. 
For example, the post (a) in Section\ \ref{sec:motiv} is annotated as $1$ because it casts suspicion on the article.
By contrast, the post (b) is annotated as $-1$ because it is regarded as the one that just mentions personal impression.
The upper part of Table~\ref{datastat} indicates the statistics of this dataset.
The number of samples are $7,775$, in which $1,036$ are positive and $6,739$ are negative samples.

\begin{table}[t]
    \small
	\centering
	\begin{tabular}{lr} \toprule
	Suspicion Casting Post Dataset\\ \hline
    \# Samples (pos / neg) & 7,775 (1,036 / 6,739) \\ 
    Avg. Length of Comments & 56.6 \\ \hline \hline
    Suspicious Article Dataset \\ \hline
    \# Samples (pos / neg) & 1,836 (564 / 1,272) \\ 
    Avg. Length of Comments & 60.4 \\
    Avg. Tweets / Article & 2.75 \\ \toprule
	\end{tabular}
	\caption{\label{datastat} Statistics of our datasets. ``pos" and ``neg" denotes the number of positive (i.e. suspicious casting posts or suspicious articles) and negative samples, respectively.}
\end{table}

\subsection{Dataset for Suspicious Article Detection}
\label{sec:data-tbv}
First, we collected a set of the posts referring to the same article (URL).
Second, we preprocessed and annotated the posts in the same way as in the SCP dataset creation.
Finally, we annotated $1$ to the article if a set of posts referring to the article includes at least one SCP post, and $-1$ otherwise.
The value $1$ means that the article is suspicious and to be verified by human experts, and $-1$ is not.
The lower part of Table~\ref{datastat} indicates the statistics of this dataset.
The number of samples are $1,836$, in which $564$ are positive and $1,272$ are negative samples.

\section{Experiments}
\label{sec:exp}

This section provides the benchmark results on our datasets.
Since our datasets have imbalanced class distributions, we used stratified 5-fold cross-validation to keep the distributions between true and false labels consistent in the train, development and test sets.

\subsection{Experimental Setup}
\label{sec:model}

\subsection*{Models}
We built and used the five models based on the following machine learning techniques.

\begin{description}
\item[(a)] {\bf Logistic Regression (LR)}: An L1 regularized logistic regression classification model.
The hyper-parameter $C$, representing inverse of regularization strength, was set to 20.
\item[(b)] {\bf SVM}: A support vector machine classification model \cite{Cortes1995SupportvectorN,Chang2011LIBSVMAL} using the radial basis function kernel (RBF).
The penalty parameter $C$ for the error term was set to 3000.
\item[(c)] {\bf Decision Tree (DT)}: A decision tree classification model \cite{Quinlan1986InductionOD,Quinlan1989InferringDT}.
The maximum depth of the tree parameter was set to $30$.
\item[(d)] {\bf Random Forest (RF)}: A random forest classification model \cite{Breiman2001RandomF}.
The maximum depth of the tree parameter was set to $15$.
The number of features used for prediction was set to $300$.
The number of trees in the forest was set to $90$.
\item[(e)] {\bf LSTM}: A Long Short Term Memory (LSTM) network based classification model \cite{Hochreiter1997LongSM,Gers2000LearningTF}.
Every tweet is represented as a sequence of word vectors and fed to the LSTM layer whose hidden units was set to $200$.
Then the averaged hidden unit vector is fed to the output layer with softmax activation function.
The hyperparameters of this model are described in more detail in Table\ \ref{tab:hyperparams} in the Appendix Section.
\end{description}

\subsection*{Implementation Details}
Parameters of these models were set by using cross-validation on the development set.
We used the default settings for unspecified hyper-parameters.

We implemented the LR, SVM, DT and RF models using scikit-learn \cite{scikit-learn}.
As the features for these four models, we used unigram and bigram word features.
Also, we implemented the LSTM model by using Keras \cite{chollet2015keras}.
As the features for the LSTM model, we used word embeddings trained on 4.5M tweets using Word2Wec CBOW model \cite{Mikolov2013EfficientEO,Mikolov2013DistributedRO}.
The vocabulary size of the embeddings is about 80,000.
The hyper-parameters used for Word2Vec are shown in Table\ \ref{tab:word2vec} in the Appendix Section.

\subsection{Results for Suspicion Casting Post Detection}

\begin{table}[t]
    \small
	\centering
	\begin{tabular}{lccc} \toprule
	Method & Precision & Recall & Micro-F1 \\ \hline
    Logistic Regression & 0.61 & 0.51 & 0.56 \\
    SVM & 0.61 & 0.49 & 0.55 \\
    Decision Tree & 0.45 &  0.54 & 0.49 \\
    Random Forest & 0.62 & 0.37 & 0.46 \\
    LSTM & 0.48 & 0.61 & 0.54 \\ \toprule
	\end{tabular}
	\caption{Results for suspicion casting post detection.}
    \label{tb:bp}
\end{table}

Table\ \ref{tb:bp} indicates the results for suspicion casting post detection on the test set.
Overall, the logistic regression, SVM and LSTM models yielded higher F1 scores than those of the decision tree and random forest models, and achieved compititive performance with each other. 
While some previous studies reported that LSTM-based model work better than other discrete feature based models in text classification tasks similar to ours \cite{Tang2015DocumentMW,Lee2016SequentialSC}, our LSTM-based model yielded almost the same F1 scores as those of logistic regression and SVM models.
One possible explanation for it is that while LSTM requires larger size of training samples, our dataset is relatively small.



\subsection{Results for Suspicious Article Detection}
\begin{table}[t]
    \small
	\centering
	\begin{tabular}{lccc} \hline
	Method & Precision & Recall & Micro-F1 \\ \hline
    Logistic Regression & 0.74 & 0.61 & 0.67 \\ 
    SVM & 0.75 & 0.60 & 0.67 \\ 
    Decision Tree & 0.61 & 0.60 & 0.61 \\ 
    Random Forest & 0.70 & 0.51 & 0.59 \\
    LSTM & 0.60 & 0.74 & 0.66 \\ \hline
    \end{tabular}
	\caption{Results for suspicious article detection.}
    \label{tb:rank}
\end{table}

\begin{figure}[t]
\begin{center}
\includegraphics[width=\linewidth]{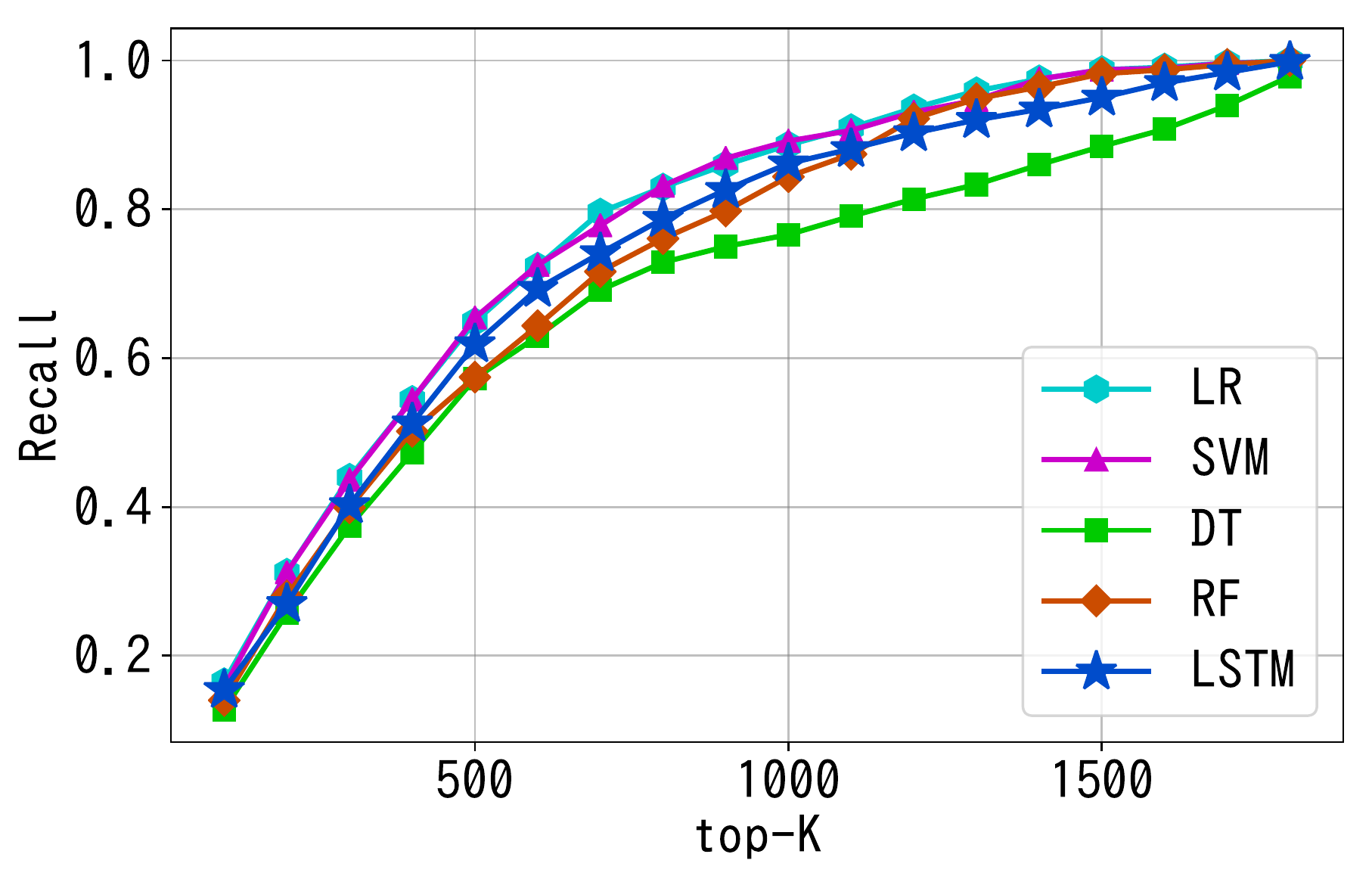}
\caption{Recall@$K$ in suspicious article detection.}
\label{fig:recall}
\end{center}
\end{figure}

Table\ \ref{tb:rank} indicates the result for suspicious article detection.
Similarily to the results in SCP detection, the logistic regression, SVM and LSTM models achieved higher scores than the other two models.

Figure\ \ref{fig:recall} shows the Recall@$K$ curve for each model.
Most of the models achieved 80\% recall at the top 750 ranked articles, which corresponds to 40\% of the total articles.
This means that by checking the top 40\% ranked articles, we can collect 80\% suspicious articles to be verified.
Thus, our models can efficiently reduce the manual cost of selecting suspicios articles.

\subsection{Analysis}
\label{sec:ana}
\subsection*{Performance Curve}
\begin{figure}[t]
\begin{center}
\includegraphics[width=\linewidth]{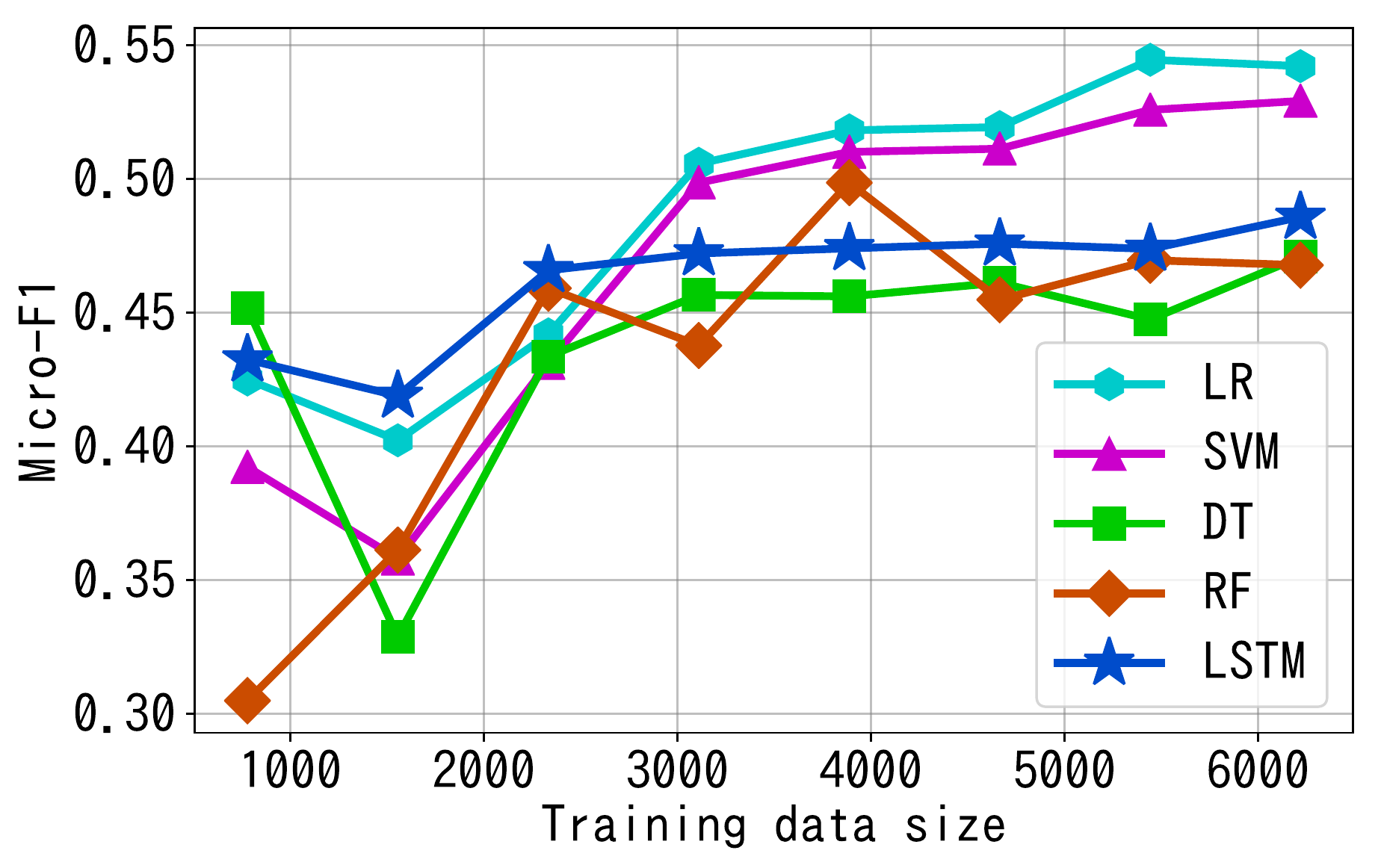}
\caption{Performance curves of each model according to the size of the training set.}
\label{fig:curve}
\end{center}
\end{figure}

\noindent
To better understand the models and benchmark results, we analyzed how the performance changes according to the size of the training set.
Figure\ \ref{fig:curve} shows the performance curve of each model.
An overall tendency we observed is that the micro-F1 scores got improved as the number of training data increased.
This result suggests that there is room for performance improvements by increasing the training data size.
As an interesting future direction, we plan to increase the data size by crowdsourcing.

\subsection*{Error Examples}
\begin{table*}[t]
	\small
    \label{ex_err}
	\begin{tabular}{lp{11.5cm}cc} \toprule
    & Tweet & Answer & Prediction\\ \hline
    \multirow{2}{*}{(1)} & これは全くの誤報、増えたのは単純労働に従事する技能実習生と留学生だろう & \multirow{2}{*}{+1} & \multirow{2}{*}{+1} \\
    & This is completely misinformation because what has increased is the number of technical intern and exchange students for manual labor. & & \\ \hline
    \multirow{2}{*}{(2)} & とうとうニュースソースきちゃったの... 誤報であって欲しかった & \multirow{2}{*}{-1} & \multirow{2}{*}{+1} \\
    & At last, the news source has got clear... I wished it had been misinformation & & \\ \hline
    \multirow{2}{*}{(3)} & 反体制派の一部に戦争犯罪があったのはかねて報道されていた通りであり、その点で記述が間違いではないのですが、戦争犯罪のレベルは天地の差があり、このタイトルはミスリード & \multirow{2}{*}{+1} & \multirow{2}{*}{-1} \\
    & As it have been reported for a long time, the description that a part of the dissidents commited a war crime is not wrong, but since the level of the war crime was so different from the reported one, this title can mislead readers. & & \\
     \toprule
	\end{tabular}
    \caption{Analysis on model predictions. The column "Answer" denotes the correct labels, and the column "Prediction" denotes the model predictions.}
\end{table*}

To shed light on the tendency of what post is difficult to predict in SCP detection, we analyze the predicted results.
Table\ \ref{ex_err} shows the examples of the predictions.

The post of example (1) points out that the article is misinformation.
All the models correctly predicted that this post is an SCP one ($+1$).
We observed that if posts contain some key phrases, such as "misinformation" and "false," the models tend to predict that they are SCP.

By contrast, all the models made wrong predictions on the post of example (2).
Like the post of example (1), this post also contains a key phrase "misinformation."
However, this post is not an SCP one ($-1$) because it just expresses the user's desire by the phrase "I wished it had been misinformation."
It is difficult for the basic models to correctly capture the meaning of the sentence-level structure.

Similar tendencies were observed in other examples.
The post of example (3) is an SCP because it denotes the title of the article can mislead readers, but all the models wrongly judged it is not an SCP.
While this post points out that the title of the article can mislead, the post also partially acknowledges the truthfulness of the content of the article by the phrase "the description ... is not wrong."
This could lead to the wrong predictions.
Since the models mainly used word-level features, it is difficult for them to properly capture sentence-level meanings.

\section{Conclusion}
\label{sec:conc}

To support human fact-checking activity, we have tackled the automation of suspicious news detection.\\

\vspace{-0.2cm}
\noindent
{\bf Summary.}\hspace{0.1cm}
To detect suspicious articles to be verified, this paper has formalized and tackled two tasks, {\it suspicion casting post detection} and {\it suspicious article detection}.
For these tasks, we have created the first publicly available dataset.
On the dataset, we have provided benchmark results using several basic machine learning techniques.
The experimental results have demonstrated that we can cover most of the suspicious articles by checking only the top ranked 40\% of the total articles.\\

\vspace{-0.2cm}
\noindent
{\bf Future Direction.} \hspace{0.1cm}
One of our future directions is to use more sophisticated models for our tasks.
Since our main objective of this work is to provide benchmark results on the datasets, we did not use complex models.
To develop systems that work well in real-world situations, it is an interesting future research to propose better models and integrate them into the systems.

Another future direction is to increase the dataset size.
As the analysis in Section~\ref{sec:ana} suggests, there is room for performance improvements by using more training samples.
We plan to increase the dataset by leveraging human experts' feedback.
In our system, human experts verify each predicted suspicious article.
In this process, we can ask the experts to correct the model predictions if they are wrong, and can add the articles and their corrected annotations (labels) to the dataset.

\section*{Acknowledgments}
This work was supported by JSPS KAKENHI Grant Number JP15H01702.We thank the anonymous reviewers for many helpful advices and comments.

\bibliography{paclic32}
\bibliographystyle{acl}

\newpage
\appendix
\section*{Appendix}
\section{Hyper-Parameters}
\label{sec:hp}

\begin{table}[h]
\label{tab:word2vec}
  \centering
  \small
  \begin{tabular}{lll} \toprule\hline
     Hyper-parameter  &  Values  \\ \hline
     Embedding size        &  300  \\
     Window size	& 7 \\
     Minimum count	&  20  \\
     Subsampling frequency	& 0.00001 \\
     Negative samples size		& 5 \\
     Epochs to train	& 5
     \\ \toprule
  \end{tabular}
  \caption{Hyper-parameters for Word2Vec training.}
\end{table}

\begin{table}[h]
\label{tab:hyperparams}
  \centering
  \small
  \begin{tabular}{lll} \toprule\hline
     Hyper-parameter  &  Values  \\ \hline
     Embedding size        &  300  \\
     Batch size            &  100 \\
     Max epoch             &  50 \\
     Optimizer			& Adam \cite{Kingma2014AdamAM} \\
     Adam $\alpha$            & \{0.002, 0.9, 0.009\}
                                    \\ \toprule
  \end{tabular}
  \caption{Hyper-parameters for the LSTM model.}
\end{table}

\end{document}